\documentclass{article}

\usepackage{microtype}
\usepackage{graphicx}
\usepackage{subcaption}
\usepackage{booktabs} 
\usepackage[pagebackref,breaklinks,colorlinks,allcolors=cvprblue]{hyperref}
\usepackage{hyperref}
\usepackage{url}
\usepackage{amsmath,bm}
\usepackage{amssymb}
\usepackage{enumitem}
\usepackage{multirow}
\usepackage{wrapfig}
\usepackage{makecell}
\usepackage[accsupp]{axessibility}
\usepackage{footnote}
\usepackage{comment}
\usepackage{arydshln} 
\usepackage[
  separate-uncertainty = true,
  multi-part-units = repeat
]{siunitx}

\usepackage[noend]{algorithmic}
\usepackage[capitalize,noabbrev]{cleveref}
\crefname{section}{Sec.}{Secs.}
\Crefname{section}{Section}{Sections}
\Crefname{table}{Table}{Tables}
\crefname{table}{Tab.}{Tabs.}

\newcommand{\ie}{\textit{i}.\textit{e}., }
\newcommand{\eg}{\textit{e}.\textit{g}., }

\usepackage[accepted]{icml2026}

\usepackage{amsmath}
\usepackage{amssymb}
\usepackage{mathtools}
\usepackage{amsthm}

\theoremstyle{plain}

\theoremstyle{definition}

\theoremstyle{remark}

\usepackage[textsize=tiny]{todonotes}

\newcommand{\nickname}{FoundObj}
\icmltitlerunning{\nickname{}: Self-supervised Foundation Models as Rewards for Label-free 3D Object Segmentation}

\begin{document}
\twocolumn
[
\icmltitle{\nickname{}: Self-supervised Foundation Models as Rewards for \\ Label-free 3D Object Segmentation}

\icmlsetsymbol{equal}{*}

\begin{icmlauthorlist}
\icmlauthor{Zihui Zhang}{equal,polyusz,vlar}
\icmlauthor{Zhixuan Sun}{equal,polyusz,vlar}
\icmlauthor{Yafei Yang}{polyusz,vlar}
\icmlauthor{Jinxi Li}{polyusz,vlar}
\icmlauthor{Jiahao Chen}{polyusz,vlar}
\icmlauthor{Bo Yang}{polyusz,vlar}
\end{icmlauthorlist}

\icmlaffiliation{polyusz}{Shenzhen Research Institute, The Hong Kong Polytechnic University;}
\icmlaffiliation{vlar}{vLAR Group, The Hong Kong Polytechnic University. \\ \phantom{xx}}
\icmlcorrespondingauthor{Bo Yang}{bo.yang@polyu.edu.hk}

\vskip 0.3in
]
\printAffiliationsAndNotice{\icmlEqualContribution}

\begin{abstract}
We address the challenging task of 3D object segmentation in complex scene point clouds without relying on any scene-level human annotations during training. Existing methods are typically constrained to identifying simple objects, primarily due to insufficient object priors in the learning process. In this paper, we present \textbf{\nickname{}}, a novel framework featuring a superpoint-based object discovery agent that incrementally merges suitable neighboring superpoints, guided by our innovative semantic and geometric reward modules. These modules synergistically leverage semantic and geometric priors from self-supervised 2D/3D foundation models, providing complementary feedback to the object discovery agent and enabling robust identification of multi-class objects through reinforcement learning. Extensive experiments on diverse benchmarks demonstrate that our approach consistently outperforms existing baselines. 
Notably, our method exhibits strong generalization in zero-shot and long-tail scenarios, underscoring its potential for scalable, label-free 3D object segmentation. 
Code is available at {\url{https://github.com/vLAR-group/FoundObj}}

\end{abstract}

\section{Introduction}

Discovering objects in 3D scenes is crucial for enabling machines to interact with the physical world, supporting a wide range of emerging applications such as autonomous driving and embodied AI. Most existing approaches \cite{Kolodiazhnyi2024,Han2025} rely heavily on dense or sparse human labels in 3D data, or on paired multi-modal data such as 2D images or text. While achieving impressive progress in closed- and open-vocabulary 3D object segmentation, these methods require substantial annotation effort, making it challenging to scale up. 
\begin{figure}[t]
\setlength{\abovecaptionskip}{ 2 pt}
\setlength{\belowcaptionskip}{ -2 pt}
\centering
\centerline{\includegraphics[width=1.\linewidth]{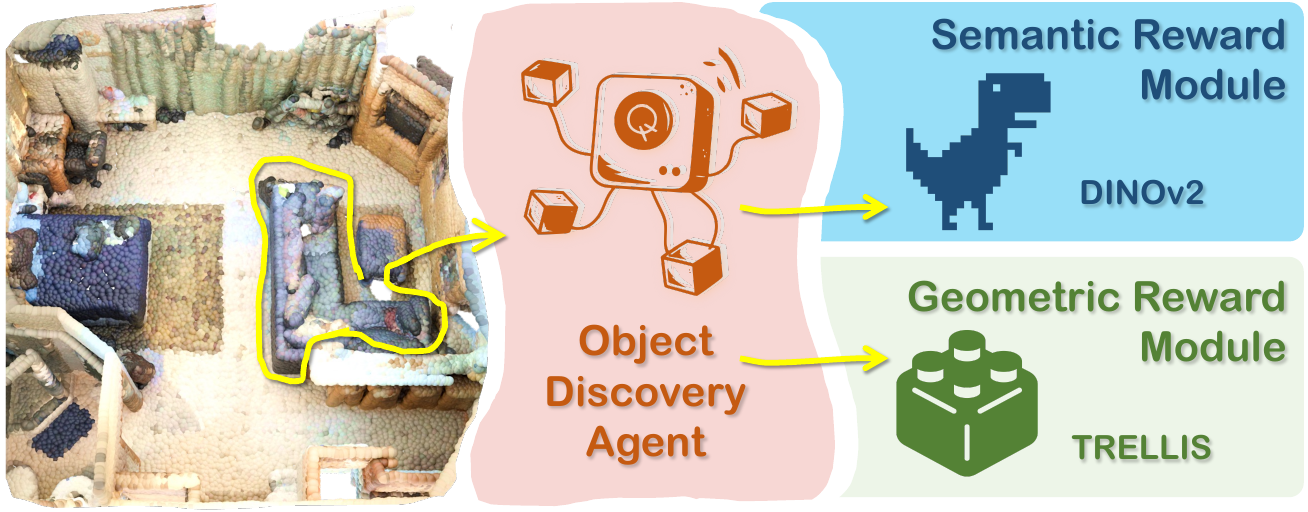}}
    \caption{Overview of our method.}
    \label{fig:meth_overview}\vspace{-0.7cm}
\end{figure}

To eliminate the dependency on manual annotations, one line of recent methods, such as UnScene3D \cite{Rozenberszki2024} and Part2Object \cite{Shi2024}, leverages self-supervised foundation models like DINO/v2 \cite{Caron2021,Oquab2024} to generate high-quality semantic features projected into 3D space for object discovery. While showing encouraging results in point clouds, they often struggle to accurately separate individual 3D objects belonging to the same category, primarily due to the absence of object geometric priors in DINO/v2. Another line of recent methods, such as EFEM \cite{Lei2023}, GrabS \cite{Zhang2025}, and its variant EvObj \cite{Chen2026}, utilizes object reconstruction models to provide fine-grained 3D geometric priors for object identification in point clouds. Despite achieving promising performance on \textit{chair} objects, they fail to discover multi-category objects with rich semantic relationships against their surroundings. 

These limitations highlight a fundamental challenge in label-free 3D object segmentation: defining what constitutes an object. Cognitive science studies \cite{Biederman1987,Chiou2016} suggest that object perception can be understood from two complementary aspects: \textit{geometry} and \textit{semantics}. Geometry characterizes object shape and structural properties, while semantics conveys the identity and meaning that distinguish one object from its surroundings. 
\begin{figure*}[t]\vspace{-0.2cm}
\setlength{\abovecaptionskip}{ 2 pt}
\setlength{\belowcaptionskip}{ -2 pt}
\centering
\centerline{\includegraphics[width=0.98\textwidth]{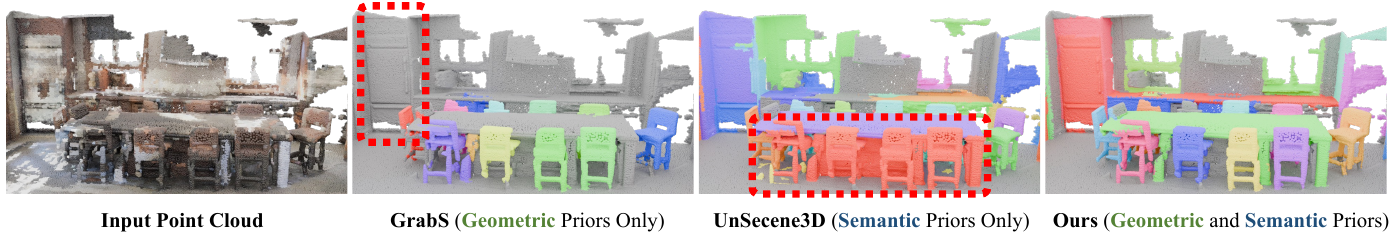}}
    \caption{Given a complex indoor 3D scene, our method can not only distinguish multiple neighboring \textit{chairs}, but also successfully identify a flat \textit{cabinet} against the wall, whereas baselines fail in one aspect or another.}
    \label{fig:opening_example}\vspace{-0.5cm}
\end{figure*}

\vspace{-0.3cm}
Building on this insight, we propose a new method for 3D object discovery that fully leverages semantic and geometric priors derived from existing self-supervised 2D/3D foundation models which have shown excellent results in various downstream tasks \cite{Gui2024,Li2024}. As illustrated in Figure \ref{fig:meth_overview}, our approach comprises three key components: (1) an object discovery agent that incrementally identifies object candidates in a spatially bottom-up manner; (2) a semantic reward module that provides feedback to the agent from existing self-supervised 2D foundation models like DINOv2 \cite{Oquab2024}; and (3) a geometric reward module that supplies feedback from 3D object-centric foundation models like TRELLIS \cite{Xiang2025}.

For the object discovery agent, given an input 3D scene point cloud, it begins with a seed superpoint and expands its spatial size by selectively merging suitable neighboring superpoints. This continues until the agent is recognized as having identified a valid object candidate, as determined by the two reward modules. Our approach is broadly inspired by the recent agent-based method GrabS \cite{Zhang2025}, which utilizes a dynamic cylinder as the agent but is limited to discovering single-class objects. In contrast, our method employs a superpoint-based agent that discovers 3D objects in a bottom-up manner, enabling the identification of objects with diverse spatial scales and structures.

The semantic and geometric reward modules are designed to provide complementary feedback to the object discovery agent, returning positive rewards when the merged superpoints are likely to form a valid object according to semantic and geometric priors, and negative rewards otherwise. To achieve this, the semantic reward module employs a new semantic consistency cut approach, ensuring that object candidates, which are exhibiting consistent semantic representations relative to their surroundings, receive positive rewards. Meanwhile, the geometric reward module utilizes a novel geometric center consistency verification mechanism, granting positive rewards to object candidates whose geometric centers demonstrate coherence. These two modules together allow us to discover multi-class 3D objects in complex point clouds through reinforcement learning (RL), without requiring human annotations during training. 

Figure \ref{fig:opening_example} shows qualitative results from an indoor 3D scene. By leveraging the semantic and geometric priors from powerful \textbf{found}ation models as rewards, our method, named \textbf{\nickname{}}, accurately discovers 3D \textbf{obj}ects, offering a distinct advantage over approaches that rely solely on semantic or geometric priors. It not only effectively separates similar objects (\eg{}\textit{chairs}) within the same semantic class, but also successfully discovers semantically complex objects (\eg{}\textit{a cabinet} on the wall) that are often overlooked by baseline methods. Our main contributions are:
\begin{itemize}[leftmargin=*]\vspace{-0.2cm}
\setlength{\itemsep}{1pt}
\setlength{\parsep}{1pt}
\setlength{\parskip}{1pt}
    \item We propose a new superpoint-based agent to discover objects by expanding their spatial sizes in a bottom-up manner, enabling the identification of diverse object shapes.     
    \item We introduce semantic and geometric reward modules that leverage rich priors from powerful foundation models, enabling the agent to be optimized without the need for human annotations in training.
    \item We demonstrate state-of-the-art object segmentation performance across multiple 3D scene benchmarks, consistently surpassing all baselines.
\end{itemize}

\section{Related Works}

\textbf{3D Object Segmentation with 3D Supervision}: Thanks to per-point human annotations in 3D datasets such as ScanNet \cite{Dai2017} and S3DIS \cite{Armeni2017}, significant progress has been made in segmenting 3D objects using both bottom-up clustering methods \cite{Wang2018d,Chen2021,Han2020,Vu2022}, top-down detection approaches \cite{Yang2019d,Yi2019,Hou2019,He2021,Shin2024}, and Transformer-based methods \cite{JiahaoLu2023,Lai2023,Schult2023,Sun2023,Kolodiazhnyi2024}. To reduce annotation costs, a range of weakly supervised methods have been developed, enabling the segmentation of 3D objects with various forms of sparse supervision, including 3D bounding boxes \cite{Chibane2022,Deng2025,Tang2022,Yoo2025} and object centers \cite{Griffiths2020}. Although these methods achieve strong performance on public benchmarks, they rely heavily on expensive human annotations, which limit their scalability in practical 3D applications.
\begin{figure*}[t]\vspace{-0.2cm}
\setlength{\abovecaptionskip}{ 2 pt}
\setlength{\belowcaptionskip}{ -2 pt}
\centering
\centerline{\includegraphics[width=0.98\textwidth]{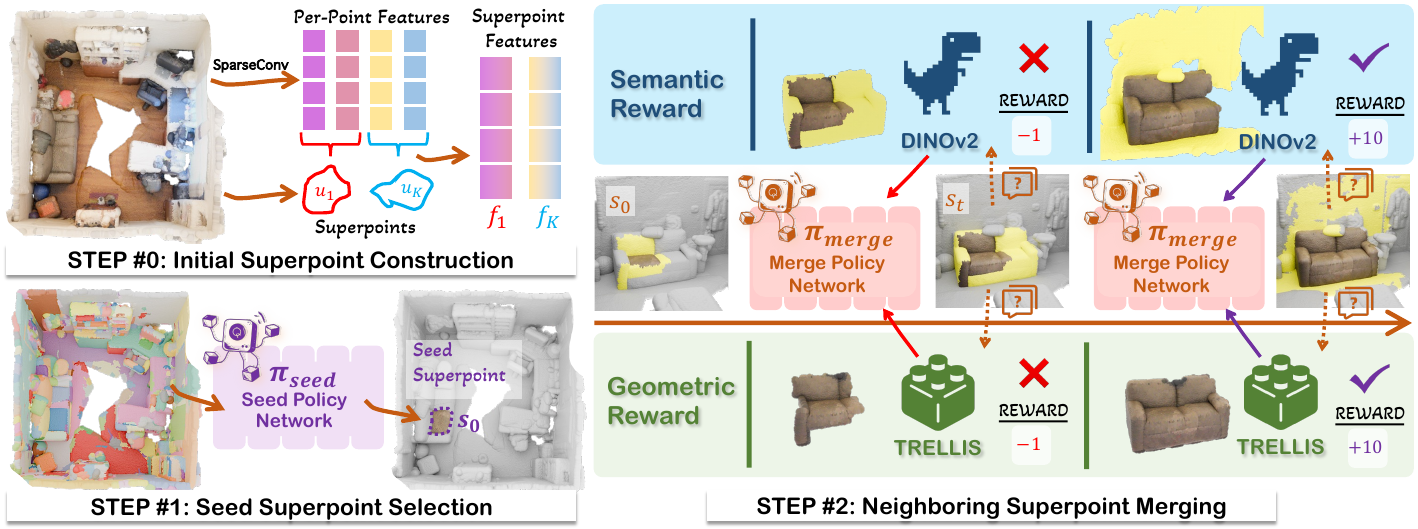}}
    \caption{Workflow of our object discovery agent. Given an input 3D scene composed of initial superpoints, our object discovery agent begins by selecting a seed superpoint and then progressively merges neighboring superpoints, guided by feedback from geometric and semantic reward modules based on self-supervised 2D/3D foundation models.}
\label{fig:meth_details}\vspace{-0.3cm}
\end{figure*}

\textbf{3D Object Segmentation with Multimodal Supervision}: With the advancement of multimodal large models such as CLIP \cite{Radford2021}, SAM \cite{Kirillov2023,Carion2025}, and LLaVA \cite{Liu2023b}, numerous subsequent methods \cite{Ha2022,Takmaz2023,Liu2023,Lu2023,Guo2024,Huang2024,Nguyen2024,Roh2024,Yan2024,Yin2024,Boudjoghra2024,Nguyen2025,Jung2025,Zhao2025,Wang2025,Lee2025,Zhou2025,Liu2025,Mei2025,Huang2026,cao2023coda} have been introduced to project pretrained 2D visual and/or vision-language features into 3D space for object discovery, enabling the identification of open-vocabulary objects. While demonstrating impressive cross-modal transfer capabilities and generalization to open-world scenarios, they still rely heavily on extensive human annotations, such as image masks, captions, or aligned image-text pairs. This dependency ultimately limits their applicability in real-world scenarios where human labels are scarce or unavailable. 

\textbf{3D Object Segmentation without Supervision}: To eliminate the need for manual annotations of 3D scenes during training, one line of unsupervised methods groups 3D points using various heuristic signals, such as surface normals, colors, or motion patterns \cite{Baur2021,Song2022,Song2024,Zhang2023,Zhang2024,Ren2026}. While effective, these approaches are often limited to discovering simple objects, such as cars. Another line of methods \cite{Rozenberszki2024,Shi2024,wang2023autorecon} projects self-supervised 2D features, such as those from DINO/v2, into 3D space, followed by point grouping. Although these methods can discover objects from multiple categories, they often struggle to distinguish between similar objects within the same category due to the inherent lack of objectness in self-supervised 2D features, as also revealed in \cite{Yang2025}. More recently, works such as GrabS \cite{Zhang2025} and its variant EvObj\cite{Chen2026}, and EFEM \cite{Lei2023} have leveraged geometric priors from object-centric reconstruction or generation models to discover objects in point clouds. While showing promising results, they are typically limited to single-class objects and are unable to identify diverse object shapes in complex environments, primarily due to the absence of semantic priors in their pipelines.

\section{\nickname{}}

Our framework consists of a superpoint-based object discovery agent (Section \ref{sec:meth_agent}), together with geometric and semantic reward modules (Sections \ref{sec:meth_geo_reward}\&\ref{sec:meth_sem_reward}) which derive feedback from existing 2D/3D foundation models. The latter two reward modules provide supervision signals to optimize the agent for discovering object candidates on 3D scene point clouds without needing human labels in training. 

\subsection{Object Discovery Agent}\label{sec:meth_agent}

As illustrated by Figure \ref{fig:meth_details}, this agent aims to identify suitable regions as object candidates, which will be scored by our two reward modules. Unlike the recent work GrabS \cite{Zhang2025} which adopts a dynamic cylinder as the agent and therefore is limited to identifying simple objects, we instead introduce a new dynamic superpoint-based agent which is highly flexible to identify any irregular object shapes. This is achieved through the following steps.

\textbf{Step \#0: Initial Superpoint Construction}: Given an input 3D scene point cloud $\bm{P}$, we first partition raw points into $K$ initial superpoints, denoted by $\{\bm{u}_1\cdot\cdot\bm{u}_k\cdot\cdot\bm{u}_K\}$ via Felzenswalb algorithm \cite{Felzenszwalb2004}. These small-sized superpoints are compact representations of the input 3D scene, enabling object discovery to be performed over $K$ regions rather than raw points, significantly reducing the exploration space for the agent. 

In parallel, we feed the raw point cloud $\bm{P}$ into an existing 3D backbone SparseConv \cite{Graham2018}, denoted by $\bm{g}_{bone}$ (not pre-trained), extracting per-point features. For the $K$ initial superpoints, we then average out per-point features within each superpoint, obtaining the corresponding $K$ superpoint features, denoted by $\{\bm{f}_1\cdot\cdot\bm{f}_k\cdot\cdot\bm{f}_K\}$. These initial superpoints will be selected and gradually merged into larger ones via the subsequent Steps \#1\&\#2. 

\textbf{Step \#1: Seed Superpoint Selection}: To discover object candidates in point cloud $\bm{P}$, our agent is designed to firstly select a seed superpoint out of $K$ as the starting point. In particular, we feed all superpoint features into a seed policy network $\bm{\pi}_{seed}$, which consists of self-attention blocks with an MLP layer followed by a softmax function, directly predicting a soft onehot code, denoted by $\bm{p}_{seed} \in \mathbb{R}^{K\times1}$.
\vspace{-0.2cm}
\begin{equation}
    \bm{p}_{seed} = \bm{\pi}_{seed}\big([\bm{f}_1\cdot\cdot\bm{f}_k \cdot\cdot \bm{f}_K]\big)
\end{equation}
The actual seed superpoint $\bm{s}_0$ is then sampled from $\bm{p}_{seed}$, and its feature vector is retrieved and denoted by $\bm{f}_0$.

\textbf{Step \#2: Neighboring Superpoint Merging}: For the seed superpoint $\bm{s}_0$, our agent then learns to select and merge some of its neighboring superpoints, getting a larger and larger superpoint which is expected to be a valid object over time. This is achieved as follows:
\begin{itemize}[leftmargin=*]\vspace{-0.3cm}
\setlength{\itemsep}{1pt}
\setlength{\parsep}{1pt}
\setlength{\parskip}{1pt}
\item \textit{Gathering All Neighboring Superpoints}: For the seed superpoint $\bm{s}_0$, we gather all its $Q$ neighboring superpoints, denoted by $\{\bm{s}_0^1 \cdots \bm{s}_0^q \cdots \bm{s}_0^Q\}$. For simplicity, we define neighboring superpoints as those within a minimum Euclidean distance of 0.1m. These $Q$ neighboring superpoints are a subset of the remaining $(K-1)$ superpoints in point cloud $\bm{P}$. Natually, we also retrieve the corresponding neighboring superpoint features, denoted by $\{\bm{f}_0^1 \cdots \bm{f}_0^q \cdots \bm{f}_0^Q\}$. 
\item \textit{Merging Neighboring Superpoints}: Now, our agent needs to learn which neighboring superpoints should be merged into the seed superpoint $\bm{s}_0$, such that the new superpoint is more likely to be an object candidate, \ie{} receiving higher rewards afterwards. To achieve this, we feed the seed and its neighboring superpoint features into a merge policy network $\bm{\pi}_{merge}$, which consists of self-attention blocks with an MLP layer followed by a sigmoid function, predicting the merging probability $\bm{p}_{merge}\in \mathbb{R}^{Q\times1}$ for $Q$ neighbors:
\vspace{-0.3cm}
\begin{equation}
    \bm{p}_{merge} = \bm{\pi}_{merge}\big(\bm{f}_0, [\bm{f}_0^1 \cdots \bm{f}_0^q \cdots \bm{f}_0^Q] \big)
\end{equation}
We then sample a subset of neighbors according to the learned policy $\bm{p}_{merge}$ and merge them into the seed superpoint $\bm{s}_0$, obtaining a larger superpoint which is regarded as an object candidate, denoted by $\bm{s}_1$. \vspace{-0.3cm}
\end{itemize}
This merging process is repeated for multiple rounds until the agent is terminated by the latter reward modules or reaches a predefined maximum round $T$, generating a sequence of object candidates, denoted by: $\{\bm{s}_0, \bm{s}_1 \cdots \bm{s}_t \cdots \bm{s}_T\}$. In each round, the obtained superpoint (\ie object candidate) will be fed into our geometric and semantic reward modules discussed below. Details of the backbone $\bm{g}_{bone}$, the seed and merging policy networks $\bm{\pi}_{seed}$ and $\bm{\pi}_{merge}$ are in Appendix \ref{sec:app_obj_dis_agent}.  

\subsection{Geometric Reward Module}\label{sec:meth_geo_reward}
For an object candidate $\bm{s}_t$, this module aims to verify whether it is geometrically coherent. Thanks to the advancement of object-centric foundation models for 3D object reconstruction and generation, such as TRELLIS \cite{Xiang2025} and Hunyuan3D \cite{Lai2025} pretrained on multiple large-scale 3D object datasets like ObjaverseXL \cite{Deitke2023a}, high-quality 3D object shape representations are effectively learned via VAE technique. To fully leverage these object geometry priors, we propose a new geometric center consistency verification mechanism to compute a reward for the candidate $\bm{s}_t$ as follows. 

\textbf{Learning an Object Center Field}: Since the pretrained 3D object foundation model often consists of an auto-encoder which cannot be directly used for scoring a candidate like $\bm{s}_t$, we propose to extend the pretrained foundation encoder by adding an additional object center field as a head, with inspiration from unMORE \cite{Yang2025}. 
\begin{figure}[t]
\centering
\centerline{\includegraphics[width=.9\linewidth]{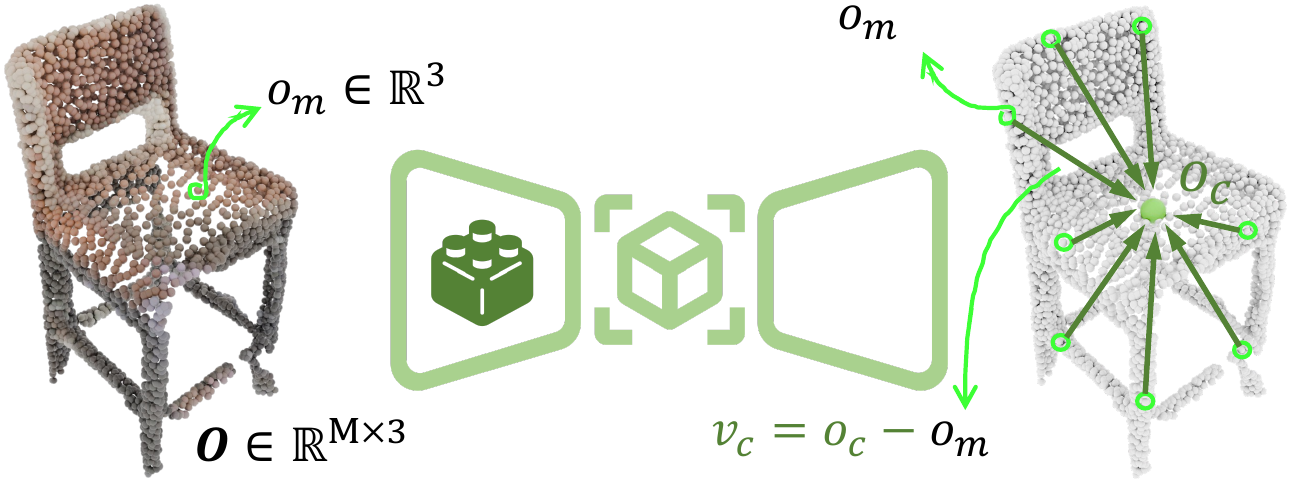}}
    \caption{An illustration of Object Center Field.}
    \label{fig:meth_obj_center_field}\vspace{-0.5cm}
\end{figure}

As illustrated in Figure \ref{fig:meth_obj_center_field}, for a 3D object $\bm{O}$ with $M$ points, denoted by $\{\bm{o}_1 \cdots \bm{o}_m \cdots \bm{o}_M\}$ and each point is represented by \textit{xyz} coordinates, its object center field is defined to indicate the direction $\bm{v}_m$ of each point pointing to the object centroid $\bm{o}_c$, mathematically as follows:
\vspace{-0.2cm}
\begin{equation}
    \bm{v}_m = \bm{o}_c - \bm{o}_m, \quad \bm{o}_c = \frac{1}{M}\sum_{m=1}^M \bm{o}_m
\end{equation}
Given the pretrained encoder from TRELLIS, we add a Transformer decoder as a head to regress the defined object center field for any query point $\bm{o}$. We train this network, denoted by $\bm{g}_{center}$, on two object datasets ABO \cite{Collins2022} and 3D-Future \cite{Fu2021} with an $\ell_2$ loss between the predicted center field and precomputed ground truth. Once well-trained, $\bm{g}_{center}$ is used to verify the geometry quality of any object candidate like $\bm{s}_t$. 

\textbf{Verifying Center Consistency}: Given a candidate $\bm{s}_t$, we directly feed it into our pretrained $\bm{g}_{center}$, estimating its corresponding center field, denoted by $\bm{v}_t$. Intuitively, if the candidate $\bm{s}_t$ is a valid object, its center field should point to a single center, meaning that $(\bm{s}_t + \bm{v}_t)$ will collapse to an extremely dense and dominant cluster. Otherwise, $(\bm{s}_t + \bm{v}_t)$ would instead have multiple or sparser clusters. 

Leveraging this property, we apply the DBSCAN clustering algorithm \cite{Ester1996} to $(\bm{s}_t + \bm{v}_t)$. If DBSCAN identifies a dominant cluster that covers at least $\alpha = 30\%$ of all points in the candidate $\bm{s}_t$ within a radius of $r = 0.05$, we assign a reward of $+10$ to the object discovery agent. Otherwise, a negative reward of $-1$ is given. Details of object center field $\bm{g}_{center}$ and training are in Appendix \ref{sec:app_geo_reward}.

\subsection{Semantic Reward Module}\label{sec:meth_sem_reward}
Geometric cues alone are often insufficient for object identification, especially in the presence of visual occlusions or cluttered backgrounds. In such cases, semantic context becomes crucial for distinguishing objects. For example, a door may be geometrically similar to a wall, but visual contrast can help delineate the boundary between them. Similarly, a chair that is largely occluded by a table may still be identified through its co-occurrence with other pieces of furniture in the scene. With this insight, this module aims to further leverage semantic priors emerging from self-supervised 2D foundation models to provide feedback for the object candidate $\bm{s}_t$.

Given a pretrained DINOv2 model, we utilize the input 3D scene point cloud $\bm{P}$ along with its associated 2D images, which are commonly available in practice. Following the approach of UnScene3D \cite{Rozenberszki2024}, we project 2D image features into 3D space using depth images. For each point in $\bm{P}$, if multiple feature vectors are projected onto it, we simply average them. The resulting point features derived from DINOv2 then serve as semantic features of the 3D scene $\bm{P}$. To calculate a reward for the object candidate $\bm{s}_t$, we propose a new semantic consistency cut approach.
\begin{figure}[t]
\centering
\centerline{\includegraphics[width=1.\linewidth]{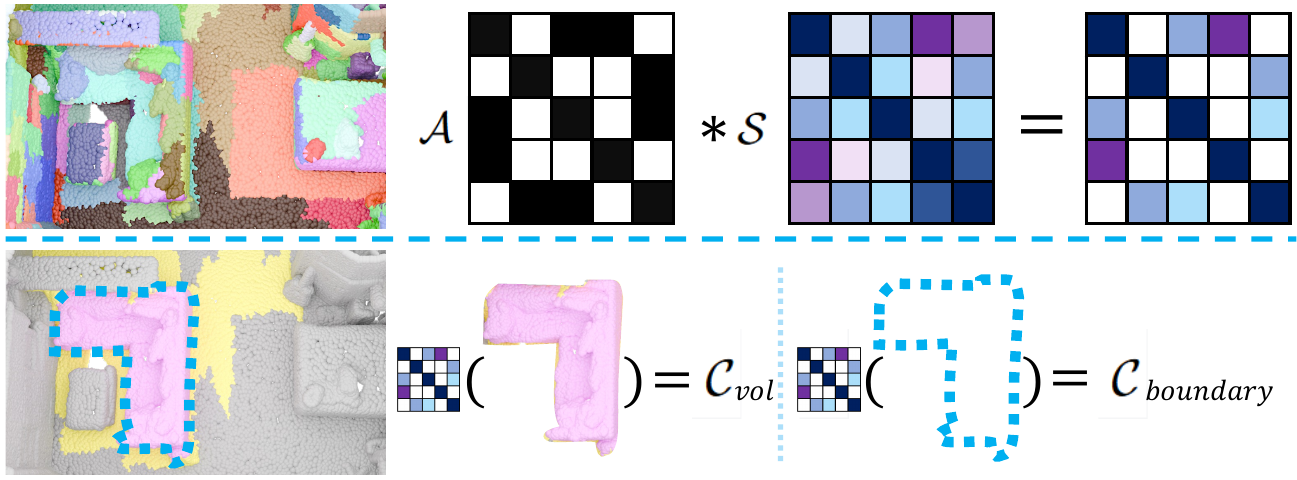}}
    \caption{An illustration of Semantic Consistency Cut.}
    \label{fig:meth_sem_cut}\vspace{-0.5cm}
\end{figure}

\textbf{Semantic Consistency Cut}: As illustrated in Figure \ref{fig:meth_sem_cut}, we assume the candidate $\bm{s}_t$ is formed by merging a total of $J$ initial superpoints over discovery as discussed in Section \ref{sec:meth_agent}, whereas the entire 3D scene point cloud $\bm{P}$ has $K$ initial superpoints. For each initial superpoint, we first compute its semantic features by averaging the projected per-point DINOv2 features. Then, we construct a pair-wise semantic similarity matrix, denoted by $\mathcal{S}\in\mathbb{R}^{K\times K}$, through calculating the cosine similarity between any two superpoints of the entire scene $\bm{P}$. In the meantime, we also construct a binary adjacency matrix, denoted by $\mathcal{A}\in\mathbb{R}^{K\times K}$, where $\mathcal{A}_{ij} =1 $ represents that the $i^{th}$ and $j^{th}$ initial superpoints are spatially adjacent, also based on a minimum Euclidean distance of 0.1m as used in Section \ref{sec:meth_agent}. Then, the resulting matrix $(\mathcal{S}*\mathcal{A})$ represents the joint spatial and semantic similarity of all initial superpoints of the 3D scene $\bm{P}$. 

To measure the semantic consistency of the object candidate $\bm{s}_t$, which can be represented by a one-hot mask $O_t\in\mathbb{R}^{K\times 1}$, inspired by NCut \cite{Shi2000}, we regard this mask as a cut against the entire 3D scene. We then calculate the cut cost as follows:
\vspace{-0.1cm}
\begin{equation}
    \mathcal{C} = \mathcal{C}_{boundary}/\mathcal{C}_{vol}
\end{equation}
where $\mathcal{C}_{boundary}$ denotes the sum of joint spatial and semantic similarity scores along the boundary of $\bm{s}_t$, whereas $\mathcal{C}_{vol}$ denotes the sum of joint similarity scores within $\bm{s}_t$. Intuitively, a higher cost $\mathcal{C}$ indicates that the candidate $\bm{s}_t$ is more similar to its spatial context, suggesting it should receive a lower reward. Otherwise, a lower cost implies that the candidate is more semantically distinct from its background, deserving a higher reward. 

In our experiments, instead of choosing a fixed cost threshold, we maintain a cost bank that stores the top 20 lowest costs for each 3D scene during training. A reward of $+10$ is given to object candidates in the bank, and $-1$ to others. More details of the cost cut calculation are in Appendix \ref{sec:app_sem_reward}.

\subsection{Training and Test}\label{sec:meth_train_test}

Given an input 3D scene point cloud, the agent continuously generates object candidates during discovery, while two reward modules assign scores based on foundational geometric and semantic priors. To fully leverage both priors, we retain the higher reward from the two modules for each candidate. During each discovery trajectory, once an object candidate receives a reward of $+10$, the agent terminates, indicating that a valid object has been discovered.

The agent is trained using the standard PPO loss. Exactly following GrabS \cite{Zhang2025}, we collect discovered object masks that receive positive rewards as pseudo labels. Lastly, we train a separate 3D object segmentation network using the Mask3D \cite{Schult2023}. For efficiency during benchmark testing, we utilize this separately trained segmentation network. More details are in \ref{sec:app_train_test}.

\section{Experiments}\label{sec:exp}

\textbf{Datasets}:
We evaluate our method on two real-world indoor benchmarks and one long-tail benchmark.
(1) \textbf{ScanNet} \citep{Dai2017} is a challenging RGB-D reconstructed dataset with heavy occlusions, sensor noise, and incomplete geometry, containing 1,201 scenes for training and 312 scenes for validation.
(2) \textbf{S3DIS} \citep{Armeni2017} is another large-scale indoor dataset with greater spatial variability, consisting of six areas that cover diverse room layouts and scene scales.
(3) \textbf{ScanNet200} \citep{Rozenberszki2022} shares the same scans as ScanNet but provides a finer-grained label space with 200 categories.
According to its official protocol, object categories are grouped into \textit{head} (66), \textit{common} (68), and \textit{tail} (66), enabling a stricter evaluation under long-tailed category distributions.

\begin{figure*}[ht]
\centering 
\centerline{\includegraphics[width=1\textwidth]{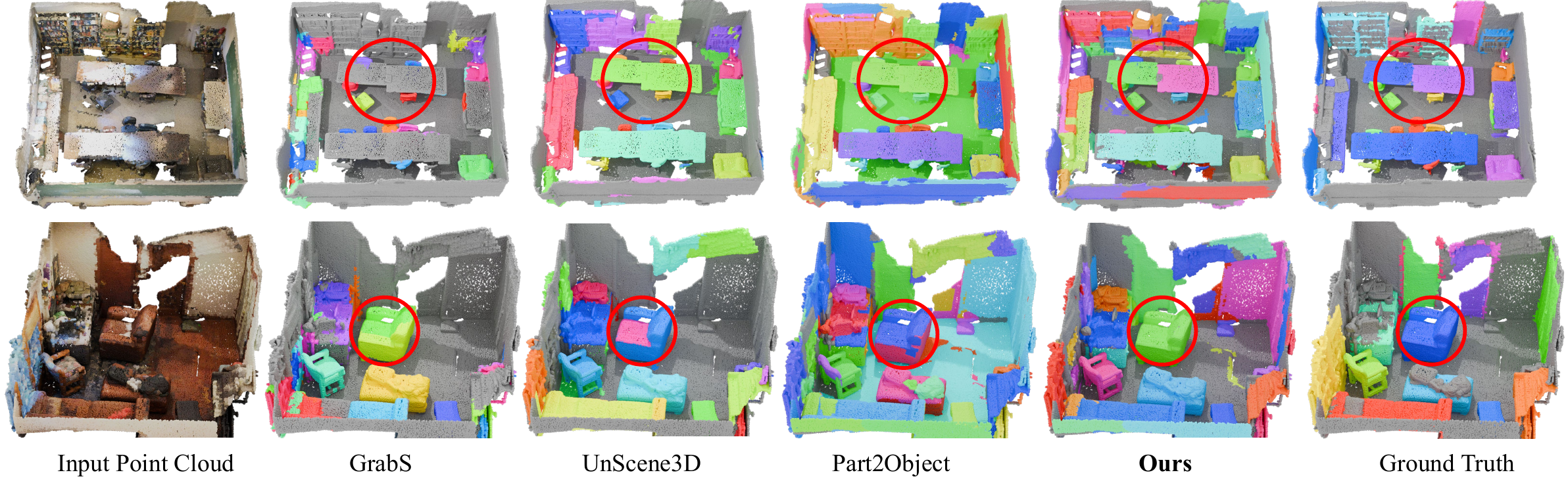}}
\vspace{-0.1cm}
    \caption{Qualitative results on the ScanNet dataset. Red circles highlight the differences.}
    \label{fig:exp_scannet}
    \vspace{-0.6cm}
\end{figure*}

\textbf{Baselines}: We compare \nickname{} with the following representative unsupervised 3D object segmentation methods that leverage either pretrained 2D priors or 3D object-centric priors.
(1) \textbf{UnScene3D} \citep{Rozenberszki2024} leverages pretrained CSC \citep{Fang2023} and DINO \citep{Caron2021} features to generate pseudo masks for training a 3D segmentation network, and we report the results using its official checkpoints of three variants.
(2) \textbf{Part2Object} \citep{Shi2024} projects pixel-level pseudo masks derived from DINOv2 features into 3D to obtain object segments.
(3) \textbf{EFEM} \citep{Lei2023} learns object priors from ShapeNet \cite{Chang2015} and performs scene-level object segmentation via an EM-style optimization procedure.
(4) \textbf{GrabS} \citep{Zhang2025} formulates unsupervised 3D object segmentation as a two-stage pipeline with an object prior network and a scene exploration agent, but the original method trains the object-prior network only on \textit{chair} objects.

\textbf{Metrics}: Following baselines, we also report class-agnostic object segmentation performance using the standard Average Precision (\textbf{AP}) protocol on ScanNet-style benchmarks \citep{Dai2017}. We report AP at IoU thresholds of 25\% (\textbf{AP@25}), 50\% (\textbf{AP@50}), and the averaged AP over IoU thresholds from 50\% to 95\% with a step size of 5\% (\textbf{AP}).

\subsection{Evaluation on ScanNet} \label{sec:exp_scannet}
We train our whole pipeline on the ScanNet training set. Following the benchmarking protocol of ScanNet, all methods are evaluated on the ScanNet validation set against ground truth object masks under the established 18-class setting. The training and validation splits are kept identical for all baselines and our \nickname{}, ensuring a fair comparison. 

Additionally, recent 3D self-supervised models such as Concerto \cite{zhang2026concerto} have shown strong capability in extracting scene-level semantics. To provide a more comprehensive evaluation, we construct additional baselines by fusing these 3D foundation model features with 2D DINOv2 features and subsequently applying the NCut algorithm as used in the UnScene3D pipeline.

\textbf{Results \& Analysis}:
Table \ref{tab:exp_scannet} and Figure \ref{fig:exp_scannet} present the quantitative and qualitative results, respectively.
Our method consistently outperforms all unsupervised baselines by a large margin. In particular, existing methods struggle to adequately segment objects, frequently missing objects or over-segmenting them into fragments.
In contrast, our \nickname{} produces more coherent and complete object masks, demonstrating the effectiveness of the geometric and semantic prior modules for object discovery in complex indoor scenes.

Compared with self-supervised baselines, our model also surpasses them by a clear margin. Notably, the fusion of TRELLIS and DINOv2 features obtains only 16.4 in AP score, as TRELLIS is trained on isolated object-level 3D data rather than 3D scenes. Therefore, its features are out-of-domain when directly applied to scene data.
Additional qualitative results are provided in Appendix \ref{sec:app_scannet}.

\begin{table}[th]\tabcolsep= 0.1cm 
\centering
 \setlength{\abovecaptionskip}{ 2 pt}
\caption{Quantitative results on 18 object categories of our method and baselines on the ScanNet validation set \cite{Dai2017}.}
\label{tab:exp_scannet}
\resizebox{1.0\linewidth}{!}{
\begin{tabular}{lccc}
\toprule[1.0pt]
\textbf{Methods} & AP & AP@50 & AP@25\\
\toprule[1.0pt]
\multicolumn{4}{l}{\textbf{Supervised:}}\\
Mask3D \cite{Schult2023} &61.2  &83.0  &93.0  \\
\midrule
\multicolumn{4}{l}{\textbf{Unsupervised:}}\\
EFEM \citep{Lei2023} &8.0  &16.7  &22.3  \\
GrabS \citep{Zhang2025} &14.0  &27.2  &39.4  \\
UnScene3D-CSC \citep{Rozenberszki2024} &16.2  &32.2  &57.6  \\
UnScene3D-DINO \citep{Rozenberszki2024} &17.7  &35.6  &62.2  \\
UnScene3D \citep{Rozenberszki2024} &18.5  &37.8  &63.7  \\
Part2Object \citep{Shi2024} &19.6  &38.4  &64.9  \\
\midrule
\multicolumn{4}{l}{\textbf{Self-supervised features followed by NCut:}}\\
Concerto & 18.2 & 38.4 & 71.6 \\
Concerto+DINOv2 & 19.8 & 41.2 & 72.2 \\
TRELLIS+DINOv2 & 16.4 & 36.8 & 66.7 \\
\midrule
\textbf{\nickname{} (Ours)} &\textbf{24.2}  &\textbf{46.2}  &\textbf{74.7}  \\
\bottomrule[1.0pt]
\end{tabular}}
\vspace{-0.5cm}
\end{table}

\subsection{Evaluation on S3DIS and ScanNet200}
\label{sec:exp_s3dis_area5}
Following the existing unsupervised methods Part2Object \citep{Shi2024} and GrabS \citep{Zhang2025}, we evaluate our method on S3DIS and ScanNet200 datasets by directly reusing our model well-trained on ScanNet, assessing the cross-dataset generalization ability.

\begin{figure*}[t]
\centering 
\centerline{\includegraphics[width=1\textwidth]{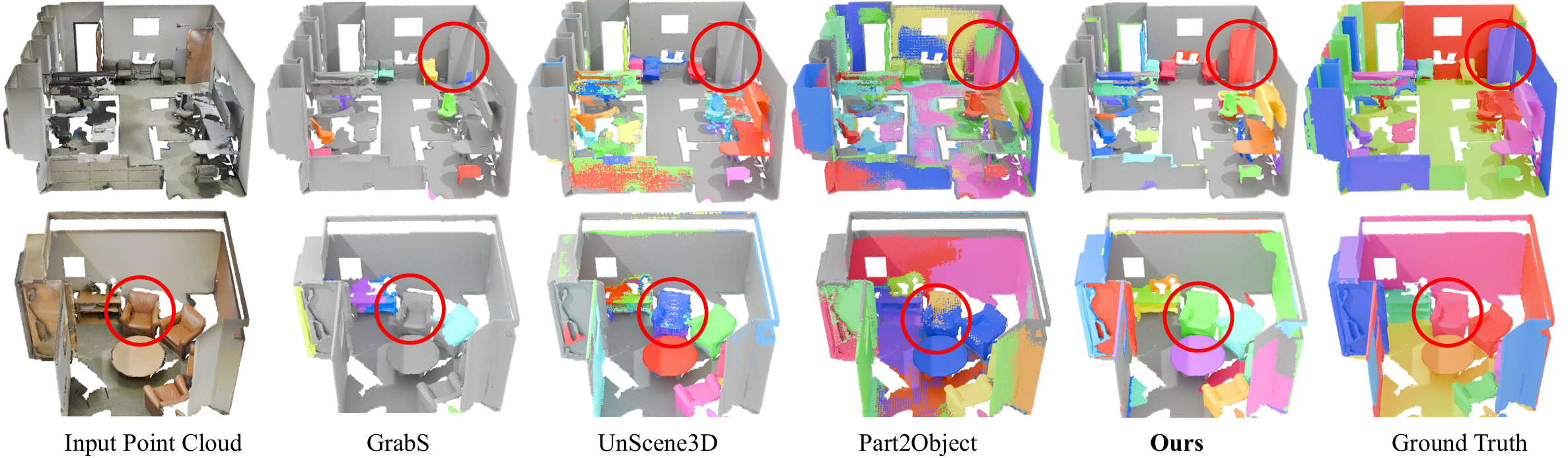}}
\vspace{-0.1cm}
    \caption{Qualitative results on the S3DIS dataset. Red circles highlight the differences.}
    \label{fig:exp_s3dis}
    \vspace{-0.1cm}
\end{figure*}

\textbf{Results on S3DIS}:
As shown in Tables \ref{tab:exp_s3dis_area5} and \ref{tab:exp_s3dis_6fold}, and  Figure \ref{fig:exp_s3dis}, our \nickname{} consistently achieves the best performance under both the Area-5 and 6-fold evaluation protocols.
These results demonstrate our strong zero-shot object segmentation capabilities, indicating that our learned object patterns generalize well across datasets with novel scene layouts. Most notably, \nickname{} achieves performance comparable to Mask3D \citep{Schult2023}, which is trained with human annotations, highlighting the significant potential of unsupervised 3D learning.

\begin{figure*}[t]
\centering 
\centerline{\includegraphics[width=1\textwidth]{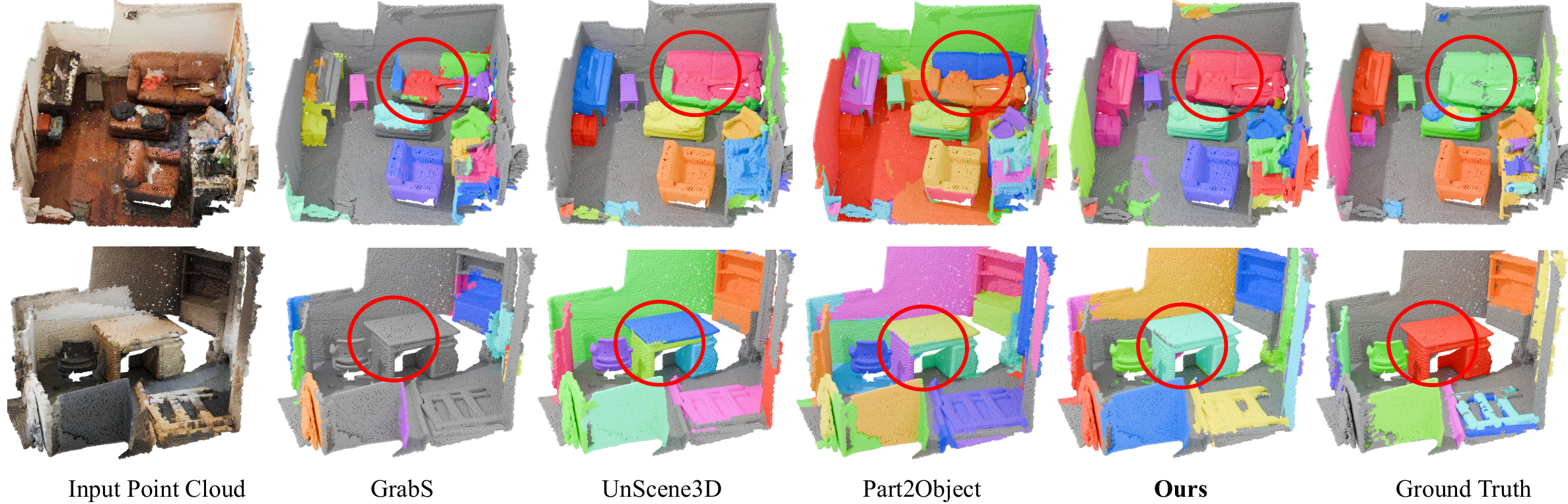}}
\vspace{-0.1cm}
    \caption{Qualitative results on the ScanNet200 dataset. Red circles highlight the differences.}
    \label{fig:exp_scannet200}
    \vspace{-0.5cm}
\end{figure*}

\textbf{Results on ScanNet200}:
On the more challenging ScanNet200 benchmark, which features a long-tailed data distribution, our method achieves clear improvements over all unsupervised baselines, as shown in Table \ref{tab:exp_scannet200} and Figure \ref{fig:exp_scannet200}.
This further demonstrates that \nickname{} is able to identify a wider variety of objects and more effectively handle long-tailed distribution.
Collectively, these cross-dataset results highlight the strong generalization ability of our method in both zero-shot and long-tail settings.
More qualitative and quantitative results are provided in Appendix \ref{sec:app_s3dis}\&\ref{sec:app_scan200}.

\begin{table}[th]\tabcolsep= 0.1cm 
\centering
 \setlength{\abovecaptionskip}{ 2 pt}
\caption{Quantitative results of our method and baselines on the S3DIS-Area5.}
\label{tab:exp_s3dis_area5}
\resizebox{1.0\linewidth}{!}{
\begin{tabular}{lccc}
\toprule[1.0pt]
\textbf{Methods} & AP & AP@50 & AP@25\\
\toprule[1.0pt]
\multicolumn{4}{l}{\textbf{Supervised:}}\\
Mask3D \cite{Schult2023} &13.0  &22.3  &37.5  \\
\midrule
\multicolumn{4}{l}{\textbf{Unsupervised:}}\\
GrabS \citep{Zhang2025} &3.7  &6.1  &9.3  \\
UnScene3D-CSC \citep{Rozenberszki2024} &8.0  &14.8  &32.2  \\
UnScene3D-DINO \citep{Rozenberszki2024} &7.0  &13.6  &32.3  \\
UnScene3D \citep{Rozenberszki2024} &8.9  &17.3  &35.9  \\
Part2Object \citep{Shi2024} &10.4  &22.5  &45.4  \\
\midrule
\textbf{\nickname{} (Ours)} &\textbf{12.8}  &\textbf{24.0}  &\textbf{45.4}  \\
\bottomrule[1.0pt]
\end{tabular}}
\vspace{-0.3cm}
\end{table}

\begin{table}[th]\tabcolsep= 0.1cm 
\centering
 \setlength{\abovecaptionskip}{ 2 pt}
\caption{Quantitative results of our method and baselines on the S3DIS 6-fold.}
\label{tab:exp_s3dis_6fold}
\resizebox{1.0\linewidth}{!}{
\begin{tabular}{lccc}
\toprule[1.0pt]
\textbf{Methods} & AP & AP@50 & AP@25\\
\toprule[1.0pt]
\multicolumn{4}{l}{\textbf{Supervised:}}\\
Mask3D \cite{Schult2023} &11.8  &20.7  &34.8  \\
\midrule
\multicolumn{4}{l}{\textbf{Unsupervised:}}\\
GrabS \citep{Zhang2025} & 3.2 & 5.5 & 9.4 \\
UnScene3D-CSC \citep{Rozenberszki2024} & 7.0 & 14.8 & 31.8 \\
UnScene3D-DINO \citep{Rozenberszki2024} & 6.2 & 14.1 & 35.5 \\
UnScene3D \citep{Rozenberszki2024} & 8.1 & 17.4 & 37.5 \\
Part2Object \citep{Shi2024} & 8.6 & 16.5 & 45.2 \\
\midrule
\textbf{\nickname{} (Ours)} & \textbf{11.4} & \textbf{24.0} & \textbf{45.7} \\
\bottomrule[1.0pt]
\end{tabular}}
\vspace{-0.3cm}
\end{table}

\begin{table}[th]\tabcolsep= 0.1cm 
\centering
 \setlength{\abovecaptionskip}{ 2 pt}
\caption{Quantitative results of our method and baselines on the ScanNet200 validation set.}
\label{tab:exp_scannet200}
\resizebox{1.0\linewidth}{!}{
\begin{tabular}{lccc}
\toprule[1.0pt]
\textbf{Methods} & AP & AP@50 & AP@25\\
\toprule[1.0pt]
\multicolumn{4}{l}{\textbf{Supervised:}}\\
Mask3D \cite{Schult2023} &26.9  &36.2  &41.4  \\
\midrule
\multicolumn{4}{l}{\textbf{Unsupervised:}}\\
EFEM \citep{Lei2023} &4.6  &9.8  &13.9  \\
GrabS \citep{Zhang2025} &7.5  &13.2  &25.6  \\
UnScene3D-CSC \citep{Rozenberszki2024} &10.3  &20.9  &42.6  \\
UnScene3D-DINO \citep{Rozenberszki2024} &11.5  &23.9  &47.3  \\
UnScene3D \citep{Rozenberszki2024} &12.8  &25.7  &49.1  \\
Part2Object \citep{Shi2024} &15.2  &31.2  &57.1  \\
\midrule
\textbf{\nickname{} (Ours)} &\textbf{18.1}  &\textbf{35.3}  &\textbf{62.8}  \\
\bottomrule[1.0pt]
\end{tabular}}
\vspace{-0.3cm}
\end{table}

\begin{table}[t]\tabcolsep= 0.05cm 
\centering
\caption{The AP scores of all ablated settings on the validation set of ScanNet based on our full \nickname{}.}
\label{tab:ablative}
\resizebox{1.0\linewidth}{!}{
\begin{tabular}{lccc}
\toprule[1.0pt]
 &AP(\%) &AP@50(\%) &AP@25(\%)\\
\toprule[1.0pt]
\textbf{Reward Modules:}\\
(1) Removing Geometric Reward Module &19.5  &40.2  &72.7 \\
(2) Removing Semantic Reward Module &15.3  &37.2  &67.6 \\
\toprule[1.0pt]
\textbf{DBSCAN Density in Geometric Reward Module:}\\
(3) $r=0.02$  &21.9  &43.6  &76.8 \\
 \textbf{(4) $r=0.05$}  &\textbf{24.2}  &\textbf{46.2}  &\textbf{74.7} \\
(5) $r=0.1$  &22.5  &43.6  &72.5 \\
(6) $\alpha=20\%$  &22.1  &43.7  &74.2 \\
\textbf{(7) $\alpha=30\%$}  &\textbf{24.2}  &\textbf{46.2}  &\textbf{74.7} \\
(8) $\alpha=40\%$  &21.8  &44.2  &75.4 \\
\toprule[1.0pt]
\textbf{Threshold for Identifying Neighboring Superpoints:}\\
(9) $d=0.05$ &23.0  &45.9  &74.8 \\
\textbf{(10) $d=0.1$} &\textbf{24.2}  &\textbf{46.2}  &\textbf{74.7} \\
(11) $d=0.2$ &21.4 &43.2 &74.9 \\
\toprule[1.0pt]
\textbf{Mask Bank Storage:}\\
(12) 10 &20.9  &41.8  &72.1 \\
\textbf{(13) 20} &\textbf{24.2}  &\textbf{46.2}  &\textbf{74.7} \\
(14) 30 &22.9 &44.7 &73.6 \\
(15) 40 &21.1 &41.4 &73.3 \\
\textbf{\nickname{} (The Full Framework)} &\textbf{24.2}  &\textbf{46.2}  &\textbf{74.7} \\
\toprule[1.0pt]
\end{tabular}
}\vspace{-0.4cm}
\end{table}

\subsection{Ablation Study}\label{exp:abl}
We conduct the following ablation studies on the ScanNet validation set to analyze the effectiveness of each component in \nickname{}, with results summarized in Table~\ref{tab:ablative}.

\textbf{- Effect of Geometric and Semantic Reward Modules}:
We evaluate the impact of the two reward modules. In particular, we either \textbf{1) remove the geometric reward module} or \textbf{2) remove the semantic reward module} to optimize our object discovery agent. We can see that it leads to a substantial performance drop, indicating that both priors are essential for object identification. Notably, removing the semantic reward module results in a larger drop. We hypothesize that DINOv2 features tend to be more discriminative than 3D priors as it is trained on a much larger dataset.

\textbf{- Sensitivity to DBSCAN Density}:
We further study the sensitivity of the geometric reward to the DBSCAN parameters.
As shown in Table \ref{tab:ablative}, setting the radius to $0.05$ yields the best performance. A smaller radius makes the density threshold overly strict, making the agent less likely to identify valid objects, while a larger radius leads to the detection of many incorrect objects. Similarly, the point ratio $\alpha$ achieves the best performance at $30\%$, whereas lower or higher values weaken the geometric prior. Overall, our model is robust to variations in density.

\textbf{- Spatial Neighboring Threshold}:
We also ablate the spatial adjacency threshold $d$ used for constructing the neighboring superpoints. As shown, $d=0.1$ achieves the best performance. A stricter threshold may incorrectly separate superpoints due to occlusions, while a larger threshold can result in object masks that are not spatially connected.

\textbf{- Semantic Cost Bank Size}:
Lastly, we analyze the effect of semantic cost bank size. Storing the top $20$ lowest costs consistently yields the best results.
A smaller bank size causes the agent to repeatedly identify only salient objects, limiting exploration diversity, whereas a larger size allows lower-quality candidates to slip in.

\begin{table}[t]
\centering
 \setlength{\abovecaptionskip}{ 2 pt}
\caption{Controlled comparisons under DINO-only, TRELLIS-only, and DINO+TRELLIS settings on the ScanNet validation set.}
\label{tab:agent_necessity}
\resizebox{1\linewidth}{!}{
\begin{tabular}{lccc}
\toprule[1.0pt]
\textbf{Methods}  &AP &AP@50 &AP@25\\
\toprule[1.0pt]
\multicolumn{4}{l}{\textbf{DINO-only:}}\\
UnScene3D \cite{Rozenberszki2024} & 17.7 & 35.6 & 65.2 \\
Part2Object \cite{Shi2024} & \textbf{19.6} & 38.4 & 64.9 \\
\textbf{\nickname{} (Ours)} & 19.5 & \textbf{40.2} & \textbf{72.7} \\
\toprule[1.0pt]
\multicolumn{4}{l}{\textbf{TRELLIS-only:}}\\
UnScene3D \cite{Rozenberszki2024} & 10.1 & 24.5 & 56.3 \\
Part2Object \cite{Shi2024} & 12.6 & 29.5 & 65.1 \\
\textbf{\nickname{} (Ours)} & \textbf{15.3} & \textbf{37.2} & \textbf{67.6} \\
\toprule[1.0pt]
\multicolumn{4}{l}{\textbf{DINO+TRELLIS:}}\\
UnScene3D \cite{Rozenberszki2024} & 15.3 & 33.2 & 68.5 \\
Part2Object \cite{Shi2024} & 17.7 & 37.5 & 70.9\\
\textbf{\nickname{} (Ours)} & \textbf{24.2} & \textbf{46.2} & \textbf{74.7} \\
\toprule[1.0pt]
\end{tabular}
}\vspace{-0.5cm}
\end{table}

\subsection{Necessity of the RL-based Object Discovery Agent}

A key question is whether the improvement of \nickname{} comes merely from using additional 3D foundation models compared with baselines \eg{} Part2Object \cite{Shi2024} and UnScene3D \cite{Rozenberszki2024}, or from the proposed RL-based discovery mechanism that effectively exploits object-level priors. To answer this, we apply the 3D foundation model TRELLIS to scene-level point clouds and evaluate the baselines under three settings: using DINOv1/v2 features exclusively, using TRELLIS features exclusively, and using a concatenation of both.

As shown in Table~\ref{tab:agent_necessity}, when only DINO features are used, \nickname{} achieves performance comparable to DINO-based clustering baselines, indicating that the agent does not merely act as a simple alternative to clustering algorithms. In contrast, under the TRELLIS-only and DINO+TRELLIS settings, \nickname{} consistently outperforms the UnScene3D and Part2Object variants, demonstrating that the RL-based agent is essential for effectively leveraging 3D object-level priors.

\subsection{Pseudo Mask Quality and Error Propagation}

Since our segmentation network is trained from pseudo masks discovered by the RL agent, we further analyze the quality of these pseudo-labels and their effect on the final Mask3D training. As shown in Table~\ref{tab:pseudo_label_quality}, without any cleaning or filtering, the discovered pseudo masks achieve 13.8 in AP score on the ScanNet training set, indicating that the agent can already discover meaningful object masks before training the final segmentation network.

To quantify the impact of pseudo mask noise, for each discovered pseudo mask, we compute its IoU with the corresponding ground-truth object mask. If the IoU is higher than 50\%, we replace the pseudo mask with the matched ground-truth mask; otherwise, we discard it. We then train Mask3D from scratch using these cleaned labels. As reported in Table~\ref{tab:pseudo_label_quality}, the resulting model achieves 37.0 in AP score on ScanNet val set, which is 12.8 points higher than training with our original pseudo masks. This confirms that pseudo mask errors are indeed propagated to the final segmentation network. Nevertheless, despite using noisy pseudo masks without any cleaning, \nickname{} still substantially outperforms previous unsupervised methods.

\begin{table}[t]
\centering
 \setlength{\abovecaptionskip}{ 2 pt}
\caption{Pseudo-label quality and error propagation on ScanNet.}
\label{tab:pseudo_label_quality}
\resizebox{0.85\linewidth}{!}{
\begin{tabular}{lccc}
\toprule[1.0pt]
  &AP &AP@50 &AP@25\\
\toprule[1.0pt]
Pseudo masks & 13.8 & 28.1 & 56.6 \\
Mask3D w/ pseudo masks & 24.2 & 46.2 & 74.7 \\
Mask3D w/ filtered masks & 37.0 & 59.7 & 83.3 \\
\toprule[1.0pt]
\end{tabular}
}\vspace{-0.2cm}
\end{table}

\begin{table}[t]\tabcolsep= 0.05cm 
\centering
 \setlength{\abovecaptionskip}{ 2 pt}
\caption{Open-vocabulary instance segmentation results on the ScanNet validation set.}
\label{tab:open_vocab}
\resizebox{1\linewidth}{!}{
\begin{tabular}{lccc}
\toprule
\textbf{Methods} &AP &AP@50 &AP@25\\
\midrule
\multicolumn{4}{l}{\textbf{Supervised:}} \\
Mask3D w/ OpenScene \cite{peng2023openscene}& 11.7 & 15.2 & 17.8 \\
OpenIns3D \cite{Nguyen2024} & 23.7 & 29.4 & 32.8 \\
OpenMask3D \cite{Takmaz2023} & 15.4 & 19.9 & 23.1 \\
\midrule
\multicolumn{4}{l}{\textbf{Unsupervised:}} \\
\nickname{} (Ours) & 6.7 & 12.7 & 16.4 \\
\bottomrule
\end{tabular}
}\vspace{-0.3cm}
\end{table}

\subsection{Extended to Open-vocabulary Segmentation}

Although \nickname{} is designed for class-agnostic object discovery, it can be naturally extended to open-vocabulary 3D object segmentation by assigning the discovered object masks with vision-language features. Specifically, after training, \nickname{} predicts object masks for each 3D scene. We then extract OpenSeg \cite{ghiasi2022scaling} features for each 3D point and average the point-wise features within each predicted mask. Finally, we compute the cosine similarity between mask features and the text embeddings of candidate class names to assign a semantic label.

We evaluate this extension on the ScanNet validation set. As shown in Table~\ref{tab:open_vocab}, although there remains a gap to fully supervised open-vocabulary methods, it is worth emphasizing that they rely on fully supervised training to obtain object masks, whereas \nickname{} generates object masks without any human annotations. These results suggest that \nickname{} provides a promising label-free object mask generator for open-vocabulary 3D scene understanding.

\subsection{Analysis on Object Discovery Agent}
\label{exp:anal}

In this section, we provide a detailed analysis of the agent’s behavior. Specifically, during training, we evaluate both the number and accuracy of discovered object candidates. A discovered object is considered \emph{accurate} if its mask achieves an IoU greater than 50\% with a matched ground truth object. We further distinguish \emph{newly discovered objects}, defined as those not identified in any previous epoch, to characterize the agent’s exploration dynamics over time. All evaluations are conducted on the ScanNet training set.

As shown in Table \ref{tab:app_discovery}, the agent discovers an increasing number of objects throughout training, eventually reaching convergence. The accuracy of discovered object candidates also improves initially and gradually stabilizes at around 40\%. Meanwhile, the number of newly discovered objects decreases over time, further indicating convergence. Additionally, the accuracy of newly discovered objects declines in later epochs, suggesting that the most salient objects are identified early, while subsequent exploration targets more challenging cases. Overall, the agent demonstrates a coarse-to-fine, progressive exploration behavior, automatically discovering a diverse range of object shapes over time.

\begin{table}[h]\tabcolsep= 0.05cm 
\centering
 \setlength{\abovecaptionskip}{ 2 pt}
\caption{The number and accuracy of object candidates discovered by the agent after different training epochs.}
\resizebox{0.99\linewidth}{!}{
\begin{tabular}{lcccccc}
\toprule[1.0pt]
Epochs &50 &100 &150 &200 &250 &300\\
\toprule[1.0pt]
\multirow{1}{*} 
Number of Obj &10408 &11362 &11599 &11750 &11775 &11810\\
Accuracy of Obj (\%) &26.6 &30.5 &35.7 &37.8 &40.1 &40.3\\
Number of New Obj &10408 &4342 &2479 &2117 &1147 &1374\\
Accuracy of New Obj (\%) &26.6 &16.8 &15.6 &14.0 &12.3 &10.3\\
\bottomrule[1.0pt]
\end{tabular}
}\label{tab:app_discovery}
\vspace{-0.3cm}
\end{table}

\section{Conclusion}

In this paper, we present \nickname{}, a novel method for effectively discovering a wide variety of 3D objects from complex real-world point clouds, without requiring human-labeled 3D scenes. Our approach introduces a superpoint-based object discovery agent, which learns to select a seed superpoint and then progressively expands its spatial size by merging suitable neighboring superpoints. By leveraging powerful self-supervised 2D/3D foundation models, our agent is guided by complementary reward modules that evaluate the semantic consistency and geometric coherence of each discovered object candidate. Extensive experiments demonstrate that \nickname{} achieves state-of-the-art performance and strong generalization in zero-shot and long-tail settings, outperforming existing unsupervised methods. 
Ablation studies and agent analyses further validate the effectiveness and robustness of each component, highlighting the potential of label-free 3D object segmentation for scalable real-world applications.
Future work will explore integrating \nickname{} into 3D pipelines like RayletDF \cite{Wei2025} to enable joint, label-free segmentation and surface reconstruction of point clouds.

\clearpage
\textbf{Acknowledgments}: This work was supported in part by Research Grants Council of Hong Kong under Grants 15219125 \& 15225522, and in part by National Natural Science Foundation of China under Grant 62271431. 

\textbf{Impact Statements}: This paper presents work whose goal is to advance the field of machine learning. There are many potential societal consequences of our work, none of which we feel must be specifically highlighted here.

\bibliography{reference}
\bibliographystyle{icml2026}

\clearpage
\appendix

\section{Details of Object Discovery Agent}\label{sec:app_obj_dis_agent}
\textbf{Backbone Network.}
Our framework starts with a 3D scene backbone $\bm{g}_{bone}$ to extract per-point features.
We adopt the Res16UNet34C architecture from SparseConv as the backbone, which consists of four downsampling and upsampling stages to capture multi-scale geometric information.
The backbone implementation is based on the \texttt{SpConv} library \citep{spconv2022}.

\textbf{Policy Networks.}
Our framework includes two policy networks, namely the seed selection policy $\boldsymbol{\pi}_{\text{seed}}$ and the neighbor merging policy $\boldsymbol{\pi}_{\text{merge}}$.
While sharing a similar design, the two policies differ in their configurations and outputs.

The seed policy network $\boldsymbol{\pi}_{\text{seed}}$ consists of a self-attention block followed by a feed-forward network (FFN) and a classification head with softmax activation.
It takes all superpoint features within a 3D scene as input.
After self-attention and FFN updates, the superpoint features are passed through an MLP and a softmax layer to produce a probability distribution over all superpoints, indicating the likelihood of each superpoint being selected as a seed.
The hidden dimension is set to 128.

The merge policy network $\boldsymbol{\pi}_{\text{merge}}$ is composed of three self-attention blocks with FFN layers.
It takes as input the features of the current region and its neighboring superpoints.
The updated features are then fed into an MLP followed by a sigmoid activation to predict the probability of each neighboring superpoint being merged into the current region.

For both policies, we introduce a learnable value token that is concatenated with the superpoint features and jointly processed through the self-attention layers.
The updated value token is finally passed to an MLP head to regress a scalar state-value estimate.
Separate value tokens and value heads are used for $\boldsymbol{\pi}_{\text{seed}}$ and $\boldsymbol{\pi}_{\text{merge}}$.

\textbf{Reinforcement Learning Optimization.}
We adopt the Proximal Policy Optimization (PPO) algorithm to train the agent.
For each trajectory, a reward of $+10$ is assigned if the current region satisfies either the geometric or semantic verification criteria, upon which the trajectory is terminated.
Otherwise, a penalty of $-1$ is assigned.
The maximum number of steps per trajectory is set to $5$.

\textbf{Seed Range Sampling Strategy.}
Seed superpoint selection requires evaluating all superpoints in a scene, which is computationally expensive and may lead the policy to repeatedly select a few dominant superpoints.
To alleviate this issue, we randomly crop a spherical region with a radius of $1$\,m from each scene during training and restrict seed selection to the superpoints within this region.
The cropped region is re-sampled at every training epoch, promoting both efficiency and exploration diversity.

\section{Details of Geometric Reward Module}\label{sec:app_geo_reward}
\textbf{Network Architecture}: The Object Center Field Network comprises an encoder and a decoder with details as follows:

For the encoder, we adopt the 3D shape encoder proposed in the VAE of TRELLIS \citep{Xiang2025}, which consists of 13 layers of 3D convolutional layers. Input point clouds are first voxelized into a resolution of \(64^3\) grid, which is then fed into the encoder to generate a resolution of \(16^3\) voxel latent representation. Each voxel in this grid is associated with a 512-dimensional feature vector, encoding the 3D shape information. We directly load their well-trained model weights and freeze the encoder parameters.

For the decoder, a self-attention block is first employed to refine the feature representations extracted by TRELLIS. Subsequently, a cross-attention block takes the refined features as input to output center-offset vectors. Notably, arbitrary query points can be fed into this cross-attention block to obtain their corresponding predicted center-offset vectors. The position embedding for query points adopts the standard Fourier embedding. Both the self-attention and cross-attention blocks are configured with a consistent feature dimension of 512.

\textbf{Date Preparation}: The Object Center Field Network is trained on 3D object datasets: ABO \citep{Collins2022} and 3D-Future \citep{Fu2021}. 
We additionally create random non-object fragments during training and enforce zero-vector predictions for their points (\ie{} $\bm{v}_m=\mathbf{0}$), improving the discriminability and robustness of the Center Field in cluttered 3D scenes.
The data preparation pipeline for training samples is detailed as follows:

First, each object mesh from the two datasets undergoes random rotation and normalization to stay within a unit cube. To simulate real-world scenarios, we append a vertical plane mesh (simulating a wall) and a horizontal plane mesh (simulating a floor) to the normalized object. Furthermore, 70\% of the training samples are augmented with additional object meshes sampled randomly from the same datasets to construct multi-object scenarios.

After creating the object meshes, we randomly select 12 views to render depth maps. The camera pitch angle for these views ranges from \(-30^\circ\) to \(+30^\circ\), and the camera is positioned 2 units away from the origin. From the rendered depth maps, 2–4 views are randomly chosen for reprojection into point clouds, which are then concatenated as partial object point clouds and used as the input to the Object Center Field Network during training. The supervision signal is precomputed as the offset from each point in the input object to the center of the object mesh. For the input non-object points, we simply set their supervision as all-zero vectors.

\section{Details of Semantic Reward Module}
\label{sec:app_sem_reward}

Given the per-superpoint DINOv2 features for a 3D scene, we aim to design a criterion to measure whether an arbitrary mask is semantically distinctive.
Inspired by NCut \citep{Shi2000}, we introduce a graph-cut cost as our semantic reward criterion.

Specifically, we build a weighted spatial graph over $K$ superpoints, where each node corresponds to a superpoint and is associated with a DINOv2 feature.
The affinity matrix considers both semantic similarity and spatial connectivity.
The semantic similarity matrix $\mathcal{S}\in\mathbb{R}^{K\times K}$ is computed using pair-wise cosine similarity between superpoints.
The spatial connectivity matrix $\mathcal{A}\in\mathbb{R}^{K\times K}$ encodes superpoint adjacency, where $\mathcal{A}_{ij}=1$ indicates that the $i^{th}$ and $j^{th}$ superpoints are spatially adjacent.
The final affinity matrix is computed as
\begin{equation}
\mathcal{W} = \mathcal{S} * \mathcal{A}.
\end{equation}

Given a binary mask $O_t\in\mathbb{R}^{K\times 1}$, we treat it as a candidate solution of the graph cut problem and partition the superpoints into two disjoint sets $O_t$ and $\bar{O}_t$.
We then compute the semantic cost following NCut \citep{Shi2000}:
\vspace{-0.1cm}
\begin{equation}\label{eq:cut}
\mathrm{cost}(O_t) = \frac{\mathrm{cut}(O_t,\bar{O}_t)}{\mathrm{vol}(O_t)},
\end{equation}
\vspace{-0.15cm}
where
\vspace{-0.1cm}
\begin{equation}
\mathrm{cut}(O_t,\bar{O}_t) = \sum_{i\in O_t}\sum_{j\in\bar{O}_t} \mathcal{W}_{ij},
\quad
\mathrm{vol}(O_t) = \sum_{i\in O_t}\sum_{j} \mathcal{W}_{ij}.
\end{equation}
\vspace{-0.1cm}

Here, $\mathrm{cut}(O_t,\bar{O}_t)$, denoted as $\mathcal{C}_{\text{boundary}}$, measures the semantic similarity of superpoint pairs across the boundary of $O_t$,
while $\mathrm{vol}(O_t)$, denoted as $\mathcal{C}_{\text{vol}}$, captures the internal semantic consistency within $O_t$.

In computation, the cut term can be expressed in matrix form using the affinity matrix $\mathcal{W}$ and the binary mask $O_t$:
\begin{equation}
\mathrm{cut}(O_t,\bar{O}_t) = O_t^{\top}\mathcal{W}(1-O_t),
\end{equation}
Similarly, the volume term can be calculated as:
\begin{equation}
\mathrm{vol}(O_t) = O_t^{\top}\mathcal{W}\mathbf{1},
\end{equation}
where $\mathbf{1}\in\mathbb{R}^{K\times 1}$ is an all-one vector.

Intuitively, this cost penalizes separating semantically similar superpoints across the boundary, while favoring regions that are internally consistent and well separated from their surrounding context.
Therefore, a lower cost indicates stronger semantic objectness for the candidate region and serves as a semantic prior for object discovery.

\section{Details of Training and Test}\label{sec:app_train_test}

For the PPO training of policy networks, we constrain the maximum change ratio between the previous and current policy distributions to $20\%$ to prevent unstable updates.
We employ generalized advantage estimation (GAE) instead of vanilla advantage regression, with the GAE parameter $\lambda=0.9$ and the discount factor $\gamma=0.9$.
To encourage exploration, an entropy regularization term is applied to the action distributions.
The overall loss consists of the PPO-Clip loss, the value regression loss, and the entropy loss, with corresponding coefficients set to $1$, $1$, and $0.1$, respectively.
We use the Adam optimizer with a learning rate of 1e-4 throughout training.

For training the segmentation network Mask3D \citep{Schult2023}, we collect all discovered object candidate masks from agent training as pseudo-labels. The training loss is the same as the vanilla Mask3D, which consists of a binary cross-entropy and dice loss for mask supervision, a cross-entropy loss for mask classification, and another binary cross-entropy for object-background classification, with weights of 2, 5, and 2. The voxel size of SparseConv is 2cm, and shares the same backbone as $\bm{g}_{bone}$. The optimizer is AdamW with a learning rate of 1e-4 in all training epochs. 

After training, we directly use the well-trained Mask3D to do inference. The usage of superpoints is also adopted, inspired by a line of 3D unsupervised works \cite{zhang2023growsp,zhang2025logosp,zhang2026growsp++,chen2026evobj}.

\section{Evaluation on ScanNet}\label{sec:app_scannet}
\begin{figure*}[t]
\centering
   \includegraphics[width=1.0\linewidth]{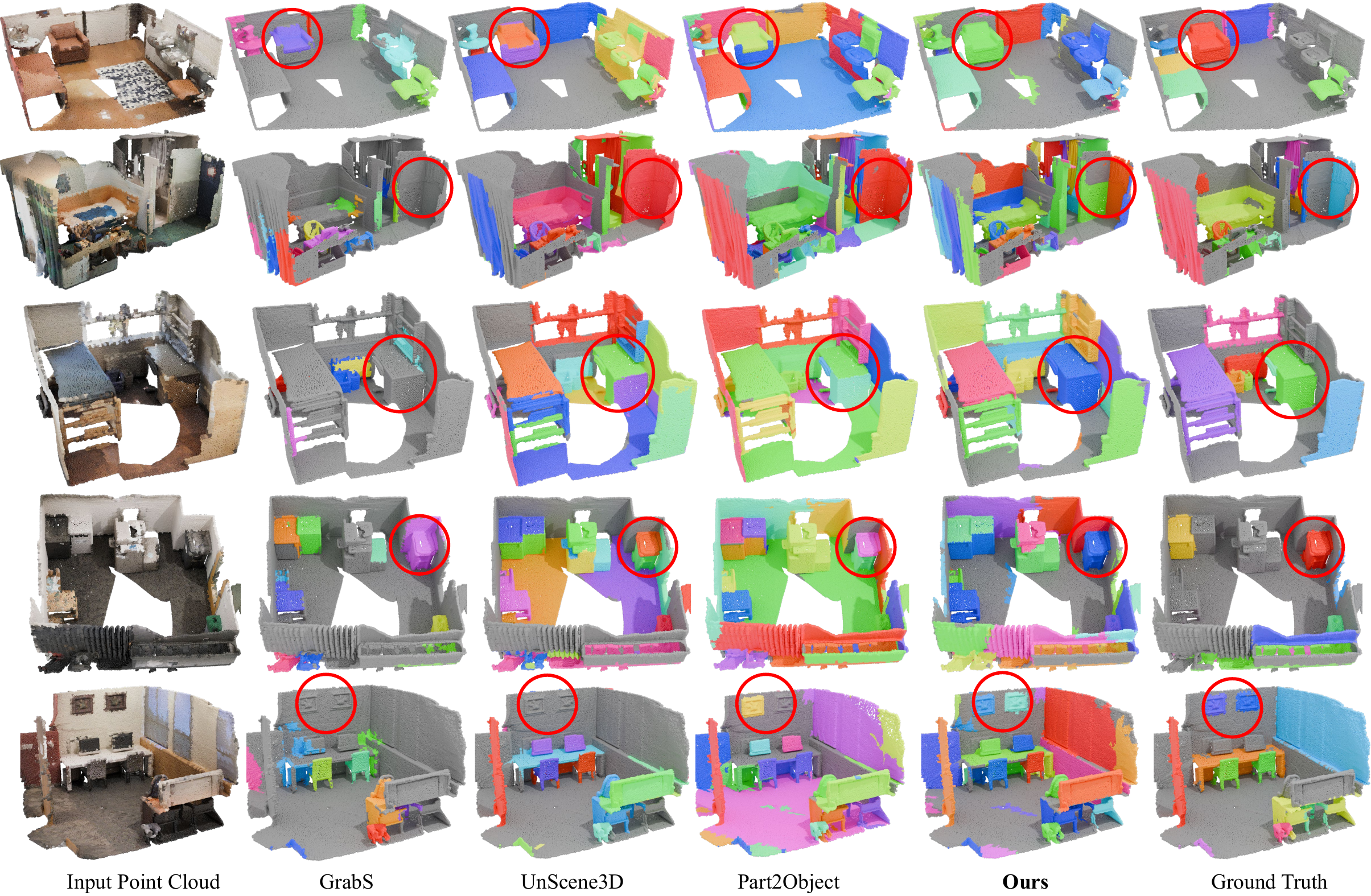}
\caption{More qualitative results on ScanNet.}
\label{fig:app_scannet}\vspace{-0.4cm}
\end{figure*}

We train our model on the ScanNet training set for 300 epochs with a batch size of 5. We use the superpoints provided by Felzenswalb algorithm \citep{Felzenszwalb2004}. Figure \ref{fig:app_scannet} provides additional qualitative comparisons with baseline methods on the ScanNet dataset. The optimizer is AdamW with a learning rate of 1e-4 in all training epochs.

\section{Evaluation on S3DIS}
\label{sec:app_s3dis}
\begin{figure*}[t]
\centering
   \includegraphics[width=1.0\linewidth]{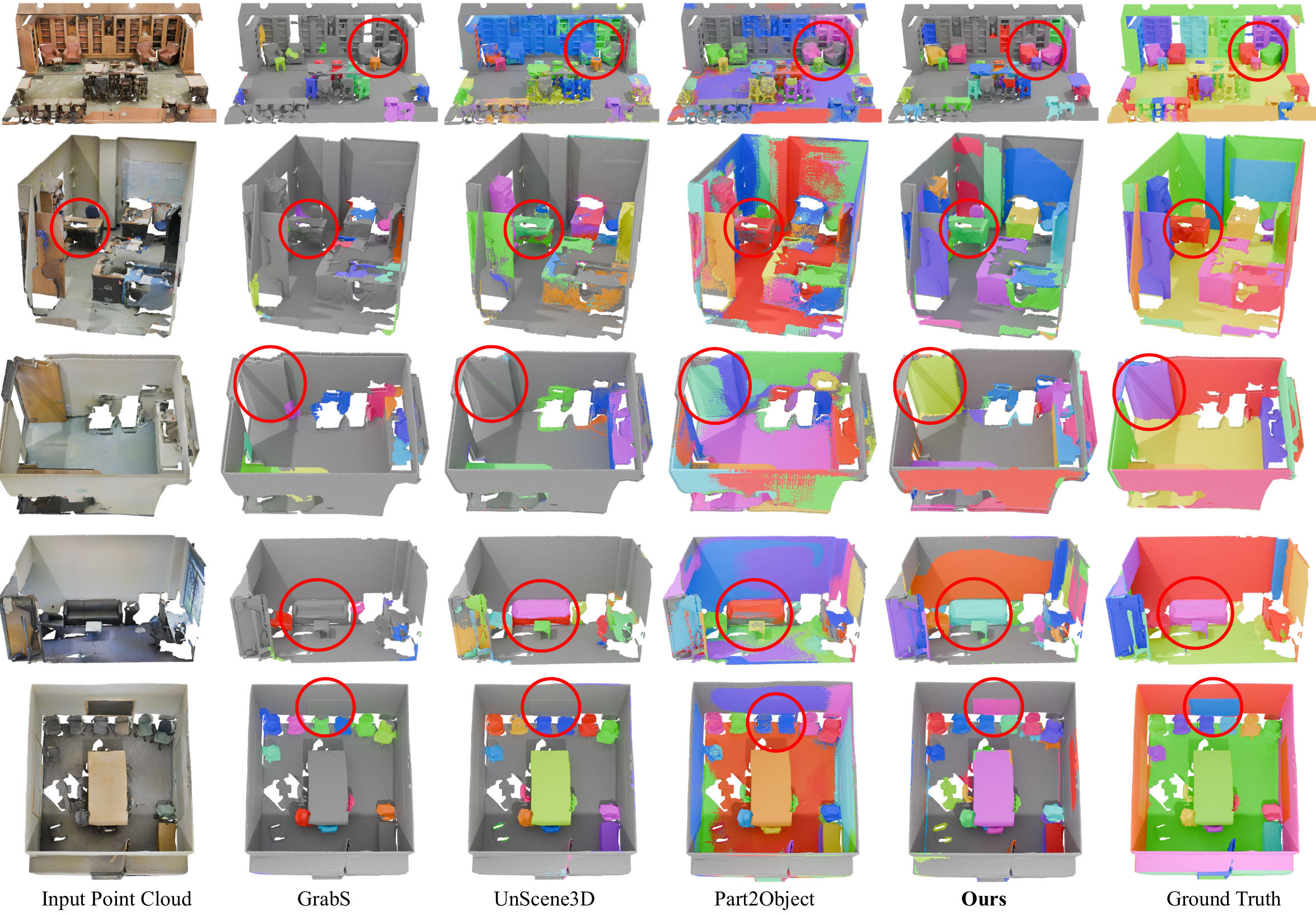}
\caption{More qualitative results on S3DIS.}
\label{fig:app_s3dis}\vspace{-0.4cm}
\end{figure*}

\cref{tab:app_exp_s3dis_area1,tab:app_exp_s3dis_area2,tab:app_exp_s3dis_area3,tab:app_exp_s3dis_area4,tab:app_exp_s3dis_area5,tab:app_exp_s3dis_area6} show the results of cross-dataset validation on each area of S3DIS. Figure \ref{fig:app_s3dis} gives more qualitative comparisons. 

\section{Evaluation on ScanNet200}
\label{sec:app_scan200}
\begin{figure*}[t]
\centering
   \includegraphics[width=1.0\linewidth]{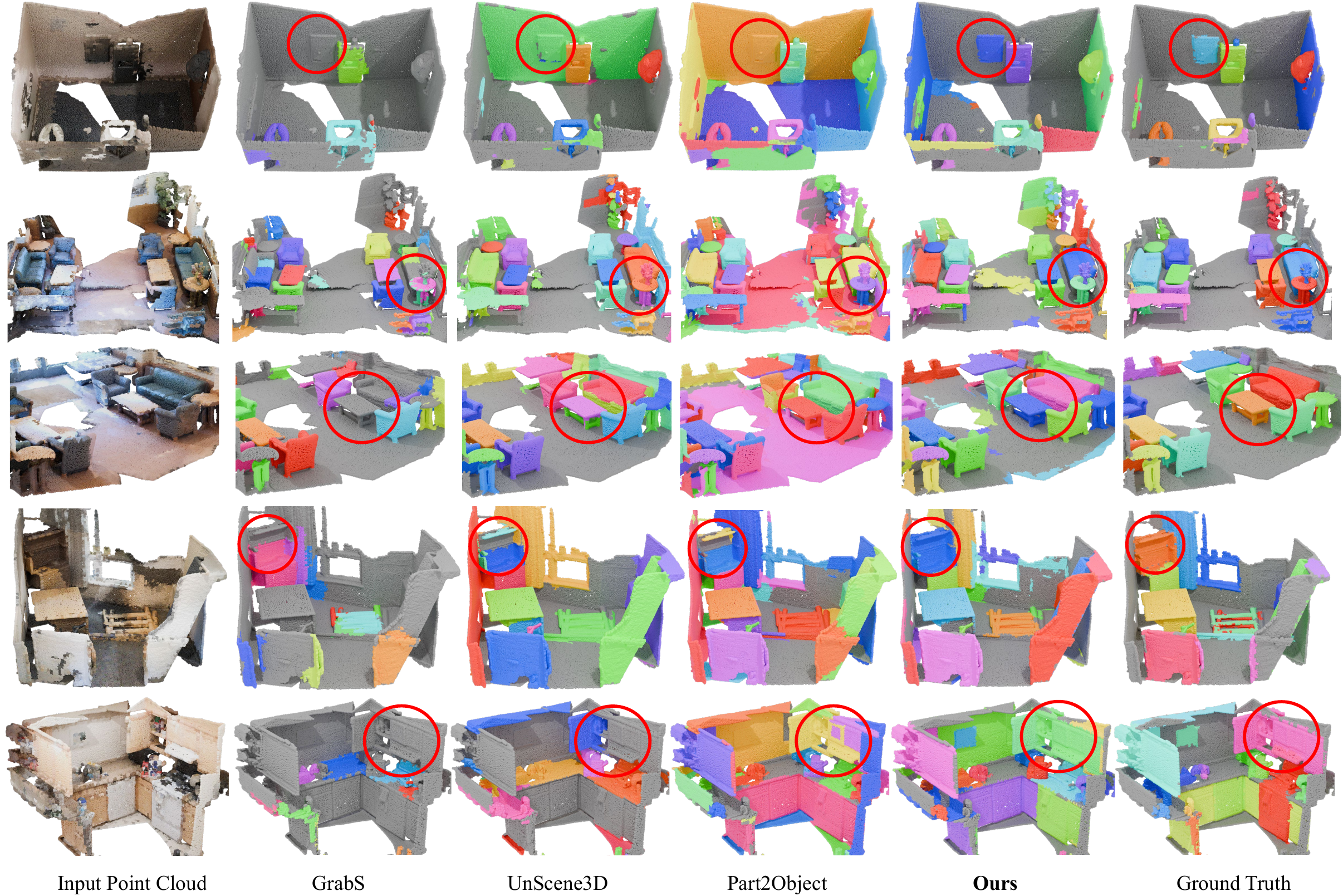}
\caption{More qualitative results on ScanNet200.}
\label{fig:app_scan200}\vspace{-0.4cm}
\end{figure*}

ScanNet200 is a more challenging benchmark; we also resume the well-trained checkpoint on ScanNet to validate the segmentation performances on this long-trial dataset. Figure \ref{fig:app_scan200} shows more qualitative results.

\section{Computational Overhead}

We also analyze the computational overhead of \nickname{}. Our framework consists of three main components. Training the Geometric Reward Module takes 13 hours and uses 16.4 GB GPU memory. The Semantic Reward Module does not require training, while extracting multi-view DINOv2 features and projecting them onto 3D point clouds takes 7 hours and 6.9 GB GPU memory. Training the object discovery agent together with the Mask3D segmentation network takes 35 hours and 14.9 GB of GPU memory. In total, \nickname{} requires 55 hours of training on a single RTX 3090 GPU with an AMD R9 7950X CPU.

For comparison, Part2Object requires 44 hours in total, including feature extraction, pseudo-label construction, and segmentation network training. UnScene3D requires 39 hours in total. Therefore, \nickname{} introduces 11 and 16 additional training hours compared with Part2Object and UnScene3D, respectively. However, this extra cost brings clear improvements of 4.6 AP, 7.8 AP@50, and 9.8 AP@25 over the strongest baseline on ScanNet. Moreover, all methods use the same Mask3D architecture at inference time, so \nickname{} has the same inference speed as the baselines, averaging 0.092 seconds per ScanNet scene. This shows that the additional computation is limited to training and does not affect deployment efficiency.

\section{More 3D Object Foundation Models}

For the geometric foundation model, we further verify that other mainstream 3D object foundation models, such as Hunyuan3D~2.0 \cite{zhao2025hunyuan3d} and Direct3D \cite{wu2024direct3d}, can also be substitutes for TRELLIS. Specifically, we use their encoders to train the Center Field module and then train the object segmentation network. The results in the attached Table \ref{tab:foundation} show that our framework is not tied to a specific foundation model and generalizes well across different choices.

\begin{table}[t]
\centering
 \setlength{\abovecaptionskip}{ 2 pt}
\caption{Segmentation performance on the ScanNet validation set with different 3D foundation models.}
\resizebox{1\linewidth}{!}
{
\begin{tabular}{clccc}
\toprule
&\textbf{3D Foundation Models} & AP & AP@50 & AP@25\\
\midrule
& Hunyuan3D 2.0 \cite{zhao2025hunyuan3d} & \textbf{24.3} & 46.1 & \textbf{75.6} \\
& Direct3D \cite{wu2024direct3d} & 22.5 & 44.9 & 76.0 \\
& TRELLIS \cite{Xiang2025} & 24.2 & \textbf{46.2} & 74.7 \\
\bottomrule[1.0pt]
\end{tabular}
}
\label{tab:foundation}
\end{table}

\begin{table}[th]\tabcolsep= 0.1cm 
\centering
 \setlength{\abovecaptionskip}{ 2 pt}
\caption{Quantitative results of our method and baselines on the S3DIS-Area1.}
\label{tab:app_exp_s3dis_area1}
\resizebox{1.0\linewidth}{!}{
\begin{tabular}{lccc}
\toprule[1.0pt]
\textbf{Methods} & AP & AP@50 & AP@25\\
\toprule[1.0pt]
\multicolumn{4}{l}{\textbf{Supervised:}}\\
Mask3D \cite{Schult2023} &10.2  &18.6  &33.8  \\
\midrule
\multicolumn{4}{l}{\textbf{Unsupervised:}}\\
GrabS \citep{Zhang2025} &3.1  &5.6  &10.5  \\
UnScene3D-CSC \citep{Rozenberszki2024} &7.9  &16.2  &36.6  \\
UnScene3D-DINO \citep{Rozenberszki2024} &6.3  &17.6  &37.6  \\
UnScene3D \citep{Rozenberszki2024} &9.0  &19.9  &40.1  \\
Part2Object \citep{Shi2024} &8.3  &20.9  &47.5  \\
\midrule
\textbf{\nickname{} (Ours)} &\textbf{11.9}  &\textbf{25.7}  &\textbf{48.0}  \\
\bottomrule[1.0pt]
\end{tabular}}
\vspace{-0.1cm}
\end{table}

\begin{table}[th]\tabcolsep= 0.1cm 
\centering
 \setlength{\abovecaptionskip}{ 2 pt}
\caption{Quantitative results of our method and baselines on the S3DIS-Area2.}
\label{tab:app_exp_s3dis_area2}
\resizebox{1.0\linewidth}{!}{
\begin{tabular}{lccc}
\toprule[1.0pt]
\textbf{Methods} & AP & AP@50 & AP@25\\
\toprule[1.0pt]
\multicolumn{4}{l}{\textbf{Supervised:}}\\
Mask3D \cite{Schult2023} &6.1  &12.4  &24.1  \\
\midrule
\multicolumn{4}{l}{\textbf{Unsupervised:}}\\
GrabS \citep{Zhang2025} &0.9  &2.0  &5.7  \\
UnScene3D-CSC \citep{Rozenberszki2024} &2.9  &8.0  &10.7  \\
UnScene3D-DINO \citep{Rozenberszki2024} &1.8  &5.6  &19.9  \\
UnScene3D \citep{Rozenberszki2024} &3.1  &7.8  &23.4  \\
Part2Object \citep{Shi2024} &4.3  &10.6  &28.3  \\
\midrule
\textbf{\nickname{} (Ours)} &\textbf{5.4}  &\textbf{12.9}  &\textbf{30.5}  \\
\bottomrule[1.0pt]
\end{tabular}}
\vspace{-0.1cm}
\end{table}

\begin{table}[th]\tabcolsep= 0.1cm 
\centering
 \setlength{\abovecaptionskip}{ 2 pt}
\caption{Quantitative results of our method and baselines on the S3DIS-Area3.}
\label{tab:app_exp_s3dis_area3}
\resizebox{1.0\linewidth}{!}{
\begin{tabular}{lccc}
\toprule[1.0pt]
\textbf{Methods} & AP & AP@50 & AP@25\\
\toprule[1.0pt]
\multicolumn{4}{l}{\textbf{Supervised:}}\\
Mask3D \cite{Schult2023} &15.2  &24.3  &40.3  \\
\midrule
\multicolumn{4}{l}{\textbf{Unsupervised:}}\\
GrabS \citep{Zhang2025} &4.8  &7.0  &10.1  \\
UnScene3D-CSC \citep{Rozenberszki2024} &8.5  &17.0  &36.9  \\
UnScene3D-DINO \citep{Rozenberszki2024} &8.0  &16.5  &38.7  \\
UnScene3D \citep{Rozenberszki2024} &9.7  &19.5  &41.9  \\
Part2Object \citep{Shi2024} &10.5  &24.7  &48.8  \\
\midrule
\textbf{\nickname{} (Ours)} &\textbf{12.6}  &\textbf{26.5}  &\textbf{51.6}  \\
\bottomrule[1.0pt]
\end{tabular}}
\vspace{-0.1cm}
\end{table}

\begin{table}[th]\tabcolsep= 0.1cm 
\centering
 \setlength{\abovecaptionskip}{ 2 pt}
\caption{Quantitative results of our method and baselines on the S3DIS-Area4.}
\label{tab:app_exp_s3dis_area4}
\resizebox{1.0\linewidth}{!}{
\begin{tabular}{lccc}
\toprule[1.0pt]
\textbf{Methods} & AP & AP@50 & AP@25\\
\toprule[1.0pt]
\multicolumn{4}{l}{\textbf{Supervised:}}\\
Mask3D \cite{Schult2023} &12.7  &22.7  &38.1  \\
\midrule
\multicolumn{4}{l}{\textbf{Unsupervised:}}\\
GrabS \citep{Zhang2025} &2.3  &4.5  &8.9  \\
UnScene3D-CSC \citep{Rozenberszki2024} &5.7  &14.0  &36.1  \\
UnScene3D-DINO \citep{Rozenberszki2024} &6.1  &14.2  &35.9  \\
UnScene3D \citep{Rozenberszki2024} &7.8  &17.8  &39.9  \\
Part2Object \citep{Shi2024} &8.2  &21.4  &48.2  \\
\midrule
\textbf{\nickname{} (Ours)} &\textbf{12.2}  &\textbf{27.5}  &\textbf{49.0}  \\
\bottomrule[1.0pt]
\end{tabular}}
\vspace{-0.1cm}
\end{table}

\begin{table}[th]\tabcolsep= 0.1cm 
\centering
 \setlength{\abovecaptionskip}{ 2 pt}
\caption{Quantitative results of our method and baselines on the S3DIS-Area5.}
\label{tab:app_exp_s3dis_area5}
\resizebox{1.0\linewidth}{!}{
\begin{tabular}{lccc}
\toprule[1.0pt]
\textbf{Methods} & AP & AP@50 & AP@25\\
\toprule[1.0pt]
\multicolumn{4}{l}{\textbf{Supervised:}}\\
Mask3D \cite{Schult2023} &13.0  &22.3  &37.5  \\
\midrule
\multicolumn{4}{l}{\textbf{Unsupervised:}}\\
GrabS \citep{Zhang2025} &3.7  &6.1  &9.3  \\
UnScene3D-CSC \citep{Rozenberszki2024} &8.0  &14.8  &32.2  \\
UnScene3D-DINO \citep{Rozenberszki2024} &7.0  &13.6  &32.3  \\
UnScene3D \citep{Rozenberszki2024} &8.9  &17.3  &35.9  \\
Part2Object \citep{Shi2024} &10.4  &22.5  &45.4  \\
\midrule
\textbf{\nickname{} (Ours)} &\textbf{12.8}  &\textbf{24.0}  &\textbf{45.4}  \\
\bottomrule[1.0pt]
\end{tabular}}
\vspace{-0.1cm}
\end{table}

\begin{table}[th]\tabcolsep= 0.1cm 
\centering
 \setlength{\abovecaptionskip}{ 2 pt}
\caption{Quantitative results of our method and baselines on the S3DIS-Area6.}
\label{tab:app_exp_s3dis_area6}
\resizebox{1.0\linewidth}{!}{
\begin{tabular}{lccc}
\toprule[1.0pt]
\textbf{Methods} & AP & AP@50 & AP@25\\
\toprule[1.0pt]
\multicolumn{4}{l}{\textbf{Supervised:}}\\
Mask3D \cite{Schult2023} &13.6  &23.9  &34.8  \\
\midrule
\multicolumn{4}{l}{\textbf{Unsupervised:}}\\
GrabS \citep{Zhang2025} &4.3  &7.5  &11.8  \\
UnScene3D-CSC \citep{Rozenberszki2024} &9.0  &18.6  &38.6  \\
UnScene3D-DINO \citep{Rozenberszki2024} &7.8  &16.8  &48.3  \\
UnScene3D \citep{Rozenberszki2024} &10.1  &22.0  &43.7  \\
Part2Object \citep{Shi2024} &9.8  &24.2  &53.0  \\
\midrule
\textbf{\nickname{} (Ours)} &\textbf{13.5}  &\textbf{27.6}  &\textbf{49.6}  \\
\bottomrule[1.0pt]
\end{tabular}}
\vspace{-0.1cm}
\end{table}

\end{document}